\documentclass[sigconf]{acmart}

 \usepackage{algorithm} 
\usepackage{algpseudocode}

\usepackage{tikz}
\usetikzlibrary{automata,positioning}
\usetikzlibrary{decorations.pathreplacing,decorations.markings,shapes.geometric}
\usetikzlibrary{shapes,arrows}
\usetikzlibrary{backgrounds,calc,positioning}

\usepackage{tabularx}       
\usepackage{float}          
\usepackage{booktabs}       

\usepackage{enumerate}

\def\be{ \begin{equation} }
\def\ee{ \end{equation} }
\def\bea{ \begin{eqnarray} }
\def\eea{ \end{eqnarray} }

\def\b0{{\bf 0}}

\catcode`,\active

\catcode`\,12

\theoremstyle{remark}

\setcopyright{none}

\acmDOI{}

\acmISBN{}

\acmPrice{}

\begin{document}
\title{Poster: Implementing Quantum Machine Learning in Qiskit for Genomic Sequence Classification}
\title{An Independent Implementation of Quantum Machine Learning Algorithms in Qiskit for Genomic Data}


\author{Navneet Singh and Shiva Raj Pokhrel}

\renewcommand{\shortauthors}{Singh and Pokhrel}

\begin{abstract}
In this paper, we explore the power of \textbf{Quantum Machine Learning} as we extend, implement and evaluate algorithms like Quantum Support Vector Classifier (QSVC), Pegasos-QSVC, Variational Quantum Circuits (VQC), and Quantum Neural Networks (QNN) in \href{https://github.com/PokhrelDeakinGitHub/Deakin_Quantum-ML_genomic}{\color{blue}Qiskit} with diverse feature mapping techniques for genomic sequence classification.\footnote{Authors are from IoT \& SE Lab, School of IT, Deakin University, Australia; shiva.pokhrel@deakin.edu.au.}

\end{abstract}
\maketitle

\section{Introduction}
The growing evolution of the Quantum Machine Learning (QML) models has shown leapfrog in medical imaging analysis, however, there remains a notable gap in genomic sequence classification, prompting our research to evaluate, rethink and extend the baseline performance of Quantum Support Vector Classifier (QSVM), Pegasos-QSVM~\cite{gentinetta2024complexity}, Variational Quantum Classifier (VQC), and Quantum Neural Network (QNN)~\cite{abbas2021power} over genomic data~\cite{pokhrel2024quantum}. 

In this preliminary research, we employ feature mapping techniques: \textit{ZFeatureMap, ZZFeatureMap, and PauliFeatureMap} \cite{ havlivcek2019supervised}, extend, implement and evaluate QML algorithms over Qiskit for genomic data, to analyze the impact of algorithmic and feature mapping parameters for better understanding and application in genomic sequence classification.

\section{Background and Implementation}

 
Feature maps facilitate QML models by translating classical data $(x_i,y_i)$ into operational quantum states~$\psi$, enabling algorithms to process and analyze information efficiently. The ZZFeatureMap uses the ZZ gate to entangle pairs of qubits, introducing phase factors based on encoded classical data and structuring interactions between qubit pairs according to the data distribution. In contrast, the ZFeatureMap employs the Z gate to introduce phase shifts to individual qubit states based on classical data, rotating qubit states accordingly. The PauliFeatureMap uses combinations of Pauli gates (X, Y, and Z) to encode classical data.


Our developed QML algorithms and Qiskit \href{https://github.com/PokhrelDeakinGitHub/Deakin_Quantum-ML_genomic}{\color{blue}{implementations}} are open source. In our QSVC (Algorithm~\ref{qsvcAlgorithm}), we begin by encoding data points $\vec{x}_i$ into quantum states $\psi$ using a tailored quantum circuit $\mathcal{C}$, and a feature map with parametrs $\theta$, $U_{\text{feature}}(\vec{x}_i, \theta)$. A quantum kernel matrix $K$ is computed, representing the inner products between $\langle \psi(\vec{x}_i), \psi(\vec{x}_j) \rangle$, which facilitates the optimization of the SVM. Training entails solving a dual quadratic programming problem (step 4, Algo. ~\ref{qsvcAlgorithm}) using classical to find the optimal separation hyperplane for labels $y_i$ in the quantum-enhanced feature space, while predictions involve preparing a $\psi$ for new data points and evaluating the decision function, $f(\vec{x})$ incorporating support vectors, kernel evaluations, and a bias term.
\begin{algorithm}[h]
\caption{Quantum Support Vector Classifier (QSVC)}
\label{qsvcAlgorithm}
\begin{algorithmic}[1]
\State Encode data $\vec{x}_i$ into quantum states $\psi$ $\leftarrow$ \;$U_{\text{feature}}(\vec{x}_i, \theta)$.
\State Construct $\mathcal{C}$ $\to$ \; Compute kernel matrix K.
\State Measure inner product, $K_{ij} \leftarrow \langle \psi(\vec{x}_i), \psi(\vec{x}_j) \rangle$.
\State Optimize support vectors and coefficients $\alpha_i$ and solve QP problem:
\[ \vec{\alpha} \leftarrow \text{solve}\left(\max \left( \sum_{i=1}^n \alpha_i - \frac{1}{2} \sum_{i,j=1}^n y_i y_j \alpha_i \alpha_j K_{ij} \right) \right) \]
subject to: $0 \leq \alpha_i \leq C$ and $\sum_{i=1}^n \alpha_i y_i = 0$.
\State Prepare $\psi$ $\leftarrow$ \;$U_{\text{feature}}(\vec{x}_i, \theta)$.

\State \textbf{return} Decision function $f(\vec{x})$ for prediction
\[
f(\vec{x}) = \text{sign}\left(\sum_{i=1}^n \alpha_i y_i K(\vec{x}, \vec{x}_i) - b\right)
\]
\end{algorithmic}
\vspace{-5 mm}
\end{algorithm}
\begin{algorithm}[h]
\caption{Pegasos-QSVC}
\label{pegasosQSVCAlgorithm}
\begin{algorithmic}[1]
\State Encode data $\vec{x}_i$ into quantum states $\psi$ $\leftarrow$ \;$U_{\text{feature}}(\vec{x}_i, \theta)$.
\State Construct $\mathcal{C}$ $\to$ \; Compute kernel matrix $K$.
\For{$t = 1$ to $T$}
\State Randomly select a subset of samples \State Compute $y_i \langle \vec{w}, \phi(\vec{x}_i) \rangle$ for each sample
\If{$y_i \langle \vec{w}, \phi(\vec{x}_i) \rangle < 1$}
 $\vec{w} \leftarrow (1 - \eta \lambda) \vec{w} + \eta y_i \phi(\vec{x}_i)$
\Else\;
 $\vec{w} \leftarrow (1 - \eta \lambda) \vec{w}$
\EndIf
\State Normalize $\vec{w}$: $\vec{w} \leftarrow \min\left(1, \frac{1/\sqrt{\lambda}}{\|\vec{w}\|}\right) \vec{w}$
\EndFor
\State Prepare $\psi$ $\leftarrow$ \;$U_{\text{feature}}(\vec{x}_i, \theta)$.
\State \textbf{return} Decision function: $f(\vec{x}) = \text{sign}(\langle \vec{w}, \phi(\vec{x}) \rangle)$.
\end{algorithmic}
\end{algorithm}

The Pegasos-QSVC, (Algorithm~\ref{pegasosQSVCAlgorithm}), in our implementation, initializes the qubits and $\theta$, including learning rate~$\eta$ and regularization factor~$\lambda$, encoding data in $\psi$ for the computation of the quantum kernel $K$. In the training loop $T$, iterative updates occur with subset evaluations, updating the weight vector $\vec{w}$ based on margin conditions and normalizing it for regularization (step 9, Algo~\ref{pegasosQSVCAlgorithm}). The predictions $f(\vec{x})$ for new data points are based on their representations of quantum states $\psi$ using the final model~$(\langle \vec{w}, \phi(\vec{x}) \rangle)$.


\begin{algorithm}[h]
\caption{Variational Quantum Classifier (VQC)}
\label{vqcAlgorithm}
\begin{algorithmic}[1]
\State Encode data $\vec{x}_i$ into quantum states $\psi$ $\leftarrow$ \;$U_{\text{feature}}(\vec{x}_i)$.
\State Construct variational circuit: parameterized gates $U(\vec{\theta})$.
\State Apply \( U(\vec{\theta}) \) to the encoded states \( \psi \).
\State Measurement $ M \leftarrow \text{measure}_{\{ |0\rangle, |1\rangle \}}(\psi)$
\State Compute the cost function $C(\vec{\theta})$.
\While{not converged}
\State Use classical optimizer: $\vec{\theta} \leftarrow \text{optimize}(C(\vec{\theta}))$
\EndWhile
\State \textbf{return} Optimized parameters: $\vec{\theta}_{\text{opt}}$.
\end{algorithmic}
\end{algorithm}

Our implementation of VQC (Algorithm~\ref{vqcAlgorithm}), begins with a qubit and $\theta$, followed by encoding into qubits~$U_{\text{feature}}(\vec{x}_i)$. A parameterized quantum circuit is constructed~$U(\vec{\theta})$, with gate $\theta$ serving as a learnable parameters~$\vec{\theta}$. Then quantum operations are applied for measuring $M$ to compute the resultant cost function~$C(\vec{\theta})$. This process is iteratively minimized by a classical optimizer adjusting the gate parameters until convergence, yielding ~$\vec{\theta}_{\text{opt}}$.

\begin{algorithm}[h]
\caption{Quantum Neural Network (QNN) Operation}
\label{qnnAlgorithm}
\begin{algorithmic}[1]
\State Encode $\vec{x}_i$ into quantum states:  $\psi$ $\leftarrow$ \;$U_{\text{feature}}(\vec{x}_i)$.
\For{each layer $l$ from $1$ to $L$}
\State Perform unitary transformation: $U_l(\theta_l)$
\EndFor
\State Perform CNOT (or custom entangling gate)
\State Measure output qubits and obtain $|\Phi\rangle$
\While{($E > \varepsilon$)}
\State Renew $\theta$ (quantum gradient descent): $\theta = \theta - \eta \nabla_{\theta} E$
\State $E = |$\textit{expected\_output $-$ \textit{measurement}$(\Phi)|^2$}
\EndWhile
\end{algorithmic}
\end{algorithm}
Our implementation of the QNN (Algorithm~\ref{qnnAlgorithm}), initializes qubits and ~$\theta$, encoding data into quantum states~$\psi$ using ~$U_{\text{feature}}(\vec{x}_i)$. Through parameterized quantum gates across layers~$l$, we process data, capture correlations of entanglement, and compute $\phi$ by quantifying qubits. Observe that, in training, a quantum version of backpropagation adjusts based on $E$ and $\eta$ until the threshold minimization is attained.

We fed the \textit{ democoding vs. intergenomic sequence dataset}~\cite{grevsova2023genomic}, containing sequences of different transcript types in our experiments using four qubits, allowing thorough evaluation of the four proposed algorithms. A subset of 100,000 genome sequences (divided into training and test sets) is selected for analysis to evaluate various QML models. Initially, the data are transformed into numerical form through text vectorization and reduced in dimensionality using PCA, then encoded into quantum data using feature mapping techniques, preparing them for processing in QML models. 
\begin{figure}[t]
\begin{center}
\includegraphics[scale=0.317]{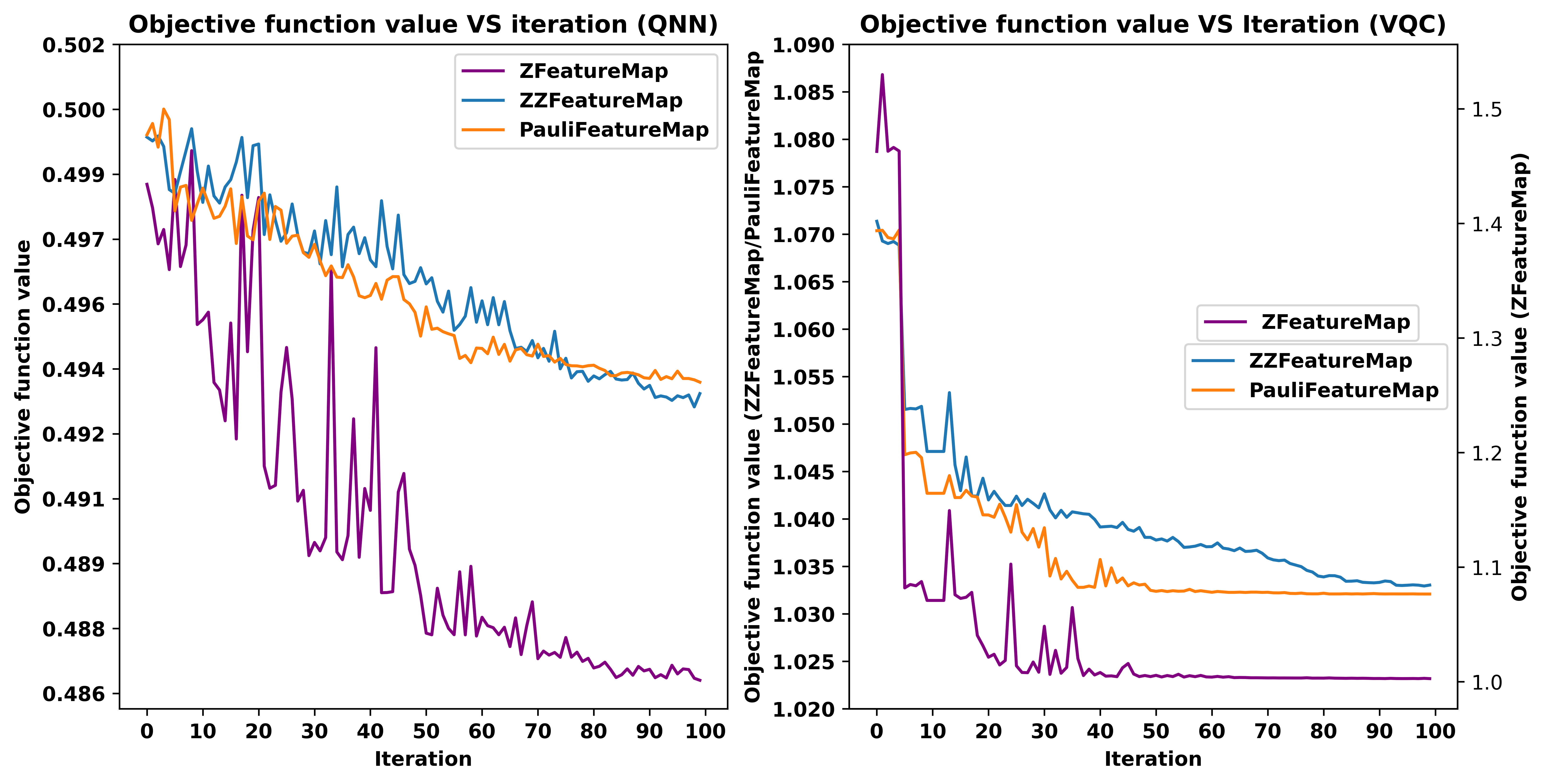}
\end{center}
\caption{\rm Convergence of QNN \& VQC objective functions.}
        \label{obj}
        \vspace{-5 mm}
\end{figure}

Figure~\ref{obj} illustrates the outperforming convergence of the objective function during the training for QNN and VQC with different feature mapping schemes (compared to~\cite{abbas2021power}). Our detailed evaluation with classification metrics is shown in Table~1, illustrating substantial improvement compared to~\cite{abbas2021power},~\cite{gentinetta2024complexity} and~\cite{pokhrel2024quantum} in accuracies (train and test), precision, recall, F1-Score, and AUROC (\textit{Area Under the Receiver Operating Characteristics}), and performance in distinguishing between the two classes of genome sequences.
\begin{figure}[!b]
    \centering
    \includegraphics[scale=0.5]{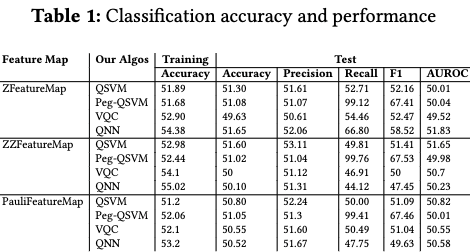}
    \label{fig:enter-label}
\end{figure}

\section{Conclusion}
Although our preliminary analysis demonstrates improved performance, ongoing detailed investigations~\cite{pokhrel2024quantum} in the realm of QML for genomic classification and evaluations with larger datasets will further illuminate the potential of these advancements.
\bibliographystyle{ACM-Reference-Format}
\bibliography{Shivaref}


\begin{thebibliography}{5}


\ifx \showCODEN    \undefined \def \showCODEN     #1{\unskip}     \fi
\ifx \showDOI      \undefined \def \showDOI       #1{#1}\fi
\ifx \showISBNx    \undefined \def \showISBNx     #1{\unskip}     \fi
\ifx \showISBNxiii \undefined \def \showISBNxiii  #1{\unskip}     \fi
\ifx \showISSN     \undefined \def \showISSN      #1{\unskip}     \fi
\ifx \showLCCN     \undefined \def \showLCCN      #1{\unskip}     \fi
\ifx \shownote     \undefined \def \shownote      #1{#1}          \fi
\ifx \showarticletitle \undefined \def \showarticletitle #1{#1}   \fi
\ifx \showURL      \undefined \def \showURL       {\relax}        \fi
\providecommand\bibfield[2]{#2}
\providecommand\bibinfo[2]{#2}
\providecommand\natexlab[1]{#1}
\providecommand\showeprint[2][]{arXiv:#2}

\bibitem[\protect\citeauthoryear{Abbas, Sutter, Zoufal, Lucchi, Figalli, and Woerner}{Abbas et~al\mbox{.}}{2021}]%
        {abbas2021power}
\bibfield{author}{\bibinfo{person}{Amira Abbas}, \bibinfo{person}{David Sutter}, \bibinfo{person}{Christa Zoufal}, \bibinfo{person}{Aur{\'e}lien Lucchi}, \bibinfo{person}{Alessio Figalli}, {and} \bibinfo{person}{Stefan Woerner}.} \bibinfo{year}{2021}\natexlab{}.
\newblock \showarticletitle{The power of quantum neural networks}.
\newblock \bibinfo{journal}{{\em Nature Computational Science\/}} \bibinfo{volume}{1}, \bibinfo{number}{6} (\bibinfo{year}{2021}), \bibinfo{pages}{403--409}.
\newblock


\bibitem[\protect\citeauthoryear{Gentinetta, Thomsen, Sutter, and Woerner}{Gentinetta et~al\mbox{.}}{2024}]%
        {gentinetta2024complexity}
\bibfield{author}{\bibinfo{person}{Gian Gentinetta}, \bibinfo{person}{Arne Thomsen}, \bibinfo{person}{David Sutter}, {and} \bibinfo{person}{Stefan Woerner}.} \bibinfo{year}{2024}\natexlab{}.
\newblock \showarticletitle{The complexity of quantum support vector machines}.
\newblock \bibinfo{journal}{{\em Quantum\/}}  \bibinfo{volume}{8} (\bibinfo{year}{2024}), \bibinfo{pages}{1225}.
\newblock


\bibitem[\protect\citeauthoryear{Gre{\v{s}}ov{\'a}, Martinek, {\v{C}}ech{\'a}k, {\v{S}}ime{\v{c}}ek, and Alexiou}{Gre{\v{s}}ov{\'a} et~al\mbox{.}}{2023}]%
        {grevsova2023genomic}
\bibfield{author}{\bibinfo{person}{Katar{\'\i}na Gre{\v{s}}ov{\'a}}, \bibinfo{person}{Vlastimil Martinek}, \bibinfo{person}{David {\v{C}}ech{\'a}k}, \bibinfo{person}{Petr {\v{S}}ime{\v{c}}ek}, {and} \bibinfo{person}{Panagiotis Alexiou}.} \bibinfo{year}{2023}\natexlab{}.
\newblock \showarticletitle{Genomic benchmarks: a collection of datasets for genomic sequence classification}.
\newblock \bibinfo{journal}{{\em BMC Genomic Data\/}} \bibinfo{volume}{24}, \bibinfo{number}{1} (\bibinfo{year}{2023}), \bibinfo{pages}{25}.
\newblock


\bibitem[\protect\citeauthoryear{Havl{\'\i}{\v{c}}ek, C{\'o}rcoles, Temme, Harrow, Kandala, Chow, and Gambetta}{Havl{\'\i}{\v{c}}ek et~al\mbox{.}}{2019}]%
        {havlivcek2019supervised}
\bibfield{author}{\bibinfo{person}{Vojt{\v{e}}ch Havl{\'\i}{\v{c}}ek}, \bibinfo{person}{Antonio~D C{\'o}rcoles}, \bibinfo{person}{Kristan Temme}, \bibinfo{person}{Aram~W Harrow}, \bibinfo{person}{Abhinav Kandala}, \bibinfo{person}{Jerry~M Chow}, {and} \bibinfo{person}{Jay~M Gambetta}.} \bibinfo{year}{2019}\natexlab{}.
\newblock \showarticletitle{Supervised learning with quantum-enhanced feature spaces}.
\newblock \bibinfo{journal}{{\em Nature\/}} \bibinfo{volume}{567}, \bibinfo{number}{7747} (\bibinfo{year}{2019}), \bibinfo{pages}{209--212}.
\newblock


\bibitem[\protect\citeauthoryear{Pokhrel, Yash, Kua, Li, and Pan}{Pokhrel et~al\mbox{.}}{2024}]%
        {pokhrel2024quantum}
\bibfield{author}{\bibinfo{person}{Shiva~Raj Pokhrel}, \bibinfo{person}{Naman Yash}, \bibinfo{person}{Jonathan Kua}, \bibinfo{person}{Gang Li}, {and} \bibinfo{person}{Lei Pan}.} \bibinfo{year}{2024}\natexlab{}.
\newblock \showarticletitle{Quantum Federated Learning Experiments in the Cloud with Data Encoding}.
\newblock \bibinfo{journal}{{\em arXiv preprint arXiv:2405.00909\/}} (\bibinfo{year}{2024}).
\newblock


\end{thebibliography}
\end{document}